\documentclass[conference]{ieeeconf}
\IEEEoverridecommandlockouts
\usepackage{graphicx}
\usepackage{cite}
\usepackage{picinpar}
\usepackage{amsmath}
\usepackage{pifont}
\usepackage{booktabs}
\usepackage{url}
\usepackage{flushend}
\usepackage[latin1]{inputenc}
\usepackage{colortbl}
\usepackage{soul}
\usepackage{multirow}
\usepackage{pifont}
\usepackage{color}
\usepackage{alltt}
\usepackage{arydshln}
\usepackage{bm}
\usepackage[hidelinks]{hyperref}
\usepackage{enumerate}
\usepackage{siunitx}
\usepackage{breakurl}
\usepackage{epstopdf}
\usepackage{pbox}
\usepackage{float}
\usepackage{balance}

\usepackage{amsmath,amssymb,amsfonts}
\usepackage{algorithmic}
\usepackage{bbding}
\usepackage{diagbox}
\usepackage{multirow}
\usepackage{subfigure}
\usepackage{multicol}
\usepackage[hidelinks]{hyperref}
\usepackage{textcomp}
\usepackage{xcolor}
\usepackage{tabularx}
\usepackage{makecell}
\usepackage{mathrsfs}
\def\BibTeX{{\rm B\kern-.05em{\sc i\kern-.025em b}\kern-.08em
    T\kern-.1667em\lower.7ex\hbox{E}\kern-.125emX}}
\begin{document}
\pagestyle{empty}
\title
{
{OpenObject-NAV: Open-Vocabulary Object-Oriented Navigation Based on Dynamic Carrier-Relationship Scene Graph}
\author{
	{Yujie Tang$^1$, Meiling Wang$^1$, Yinan Deng$^{1\dagger}$, Zibo Zheng$^{2\dagger}$, Jiagui Zhong$^{1\dagger}$, Yufeng Yue$^{1*}$}}

\thanks{
This work is  supported by the National Natural Science Foundation of China under Grant No. NSFC 62233002, 62003039.
(Corresponding Author: Yufeng Yue, yueyufeng@bit.edu.cn)}

\thanks{$^1$Yujie Tang, Meiling Wang, Yinan Deng, Jiagui Zhong, Yufeng Yue  are with
School of Automation, Beijing Institute of Technology, Beijing, 100081, China.
}
\thanks{$^2$Zibo Zheng  are with School of Mechanical Engineering, University of Nottingham Ningbo China, Ningbo, 315100, China.}

\thanks{$\dagger$: Equal contribution.}
}
 \maketitle
\begin{abstract}
In everyday life, frequently used objects like cups often have unfixed positions and multiple instances within the same category, and their carriers frequently change as well. As a result, it becomes challenging for a robot to efficiently navigate to a specific instance. To tackle this challenge, the robot must capture and update scene changes and plans continuously. However, current object navigation approaches primarily focus on semantic-level and lack the ability to dynamically update scene representation.  This paper captures the relationships between frequently used objects and their static carriers. It constructs an open-vocabulary Carrier-Relationship Scene Graph (CRSG) and updates the carrying status during robot navigation to reflect the dynamic changes of the scene. Based on the CRSG, we further propose an instance navigation strategy that models the navigation process as a Markov Decision Process. At each step, decisions are informed by Large Language Model's commonsense knowledge and visual-language feature similarity.
We designed a series of long-sequence navigation tasks for frequently used everyday items in the Habitat simulator. The results demonstrate that by updating the CRSG, the robot can efficiently navigate to moved targets. Additionally, we deployed our algorithm on a real robot and validated its practical effectiveness. The project page can be found here: \url{https://OpenObject-Nav.github.io}.
\end{abstract}


{}

\definecolor{limegreen}{rgb}{0.2, 0.8, 0.2}
\definecolor{forestgreen}{rgb}{0.13, 0.55, 0.13}
\definecolor{greenhtml}{rgb}{0.0, 0.5, 0.0}

\section{Introduction}
\label{introduction}
With the advancement of visual language models (VLM) \cite{zhang2024vision} and large language models (LLM)\cite{chang2024survey}, the realization of cognitive navigation\cite{bai2024review} has  attracted increasing attention. Imagine a daily environment, a robot is expected to navigate efficiently to any object, whether it is static furniture or a frequently used object with changing positions (such as a cup). This necessitates the robot to represent and update the current state of objects in the scene and navigate to the target. 

Current object navigation methods\cite{chang2020semantic, ye2021auxiliary,chaplot2020object,luo2022stubborn,ramakrishnan2022poni,zhang20233d, dorbala2023can, chen2023not, zhou2023esc,yokoyama2024vlfm} can effectively navigate to static objects (like a sofa). However, they are often limited to searching for semantic-level objects and lack the capability to update scenes. Hence, when it comes to frequently used daily items  such as a   ``\textit{cup on the black table},'' these objects typically come in various colors and styles. They may appear in places such as the kitchen, bedroom, and others, and their positions are not fixed. Additionally, they are usually carried by other objects, meaning that the carriers are also not fixed. Such navigation targets are highly dynamic and subject to interference, making it challenging to efficiently and effectively navigate to them.

\begin{figure}[t!]
\centering    
{
	\includegraphics[width=8cm]{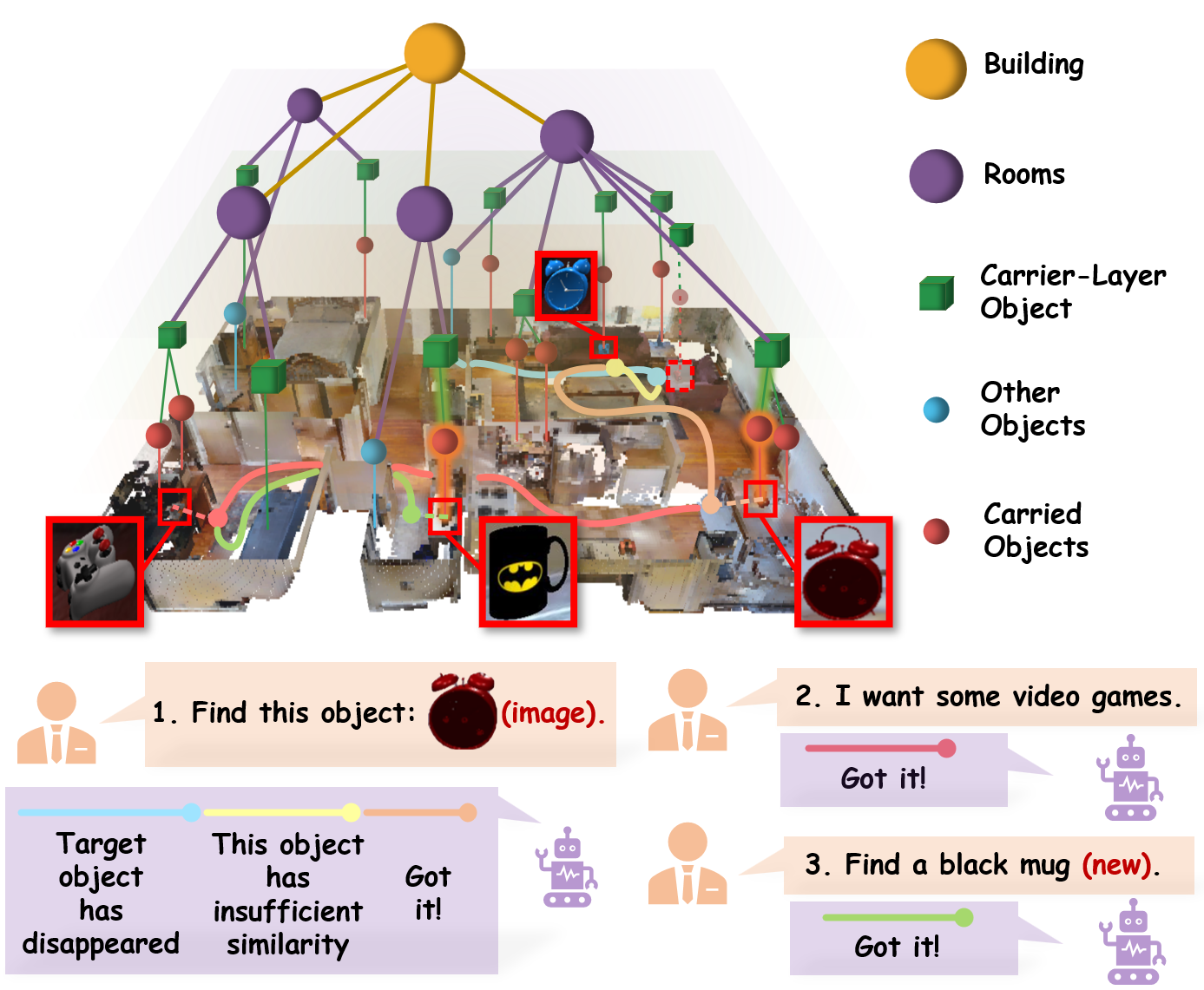} 

}
\caption{The robot executes long-sequence, multi-modal, and multi-type daily object navigation commands based on a dynamic carrier-relationship scene graph. \textbf{First}, it successfully navigates to the displaced \textit{\textbf{red alarm clock}}, eliminating interference from a \textbf{\textit{blue alarm clock}} of the same type along the way. \textbf{Next}, based on the user's request, it navigates to a \textit{\textbf{game controller}}. During these tasks, the robot observes a new \textbf{\textit{black cup}}, which is added to the scene graph. This update enables efficient point-to-point navigation for the \textbf{third} task.} %
\label{two-00} 
\end{figure} 

To appropriately represent highly dynamic and interference-prone daily environments and achieve more efficient navigation to everyday objects, \textcolor{black}{an appropriate hierarchical scene representation is essential.}
Current scene representation methods\cite{rosinol2021kimera,tian2022kimera,hughes2022hydra,chang2023hydra,hughes2024foundations,gu2024conceptgraphs,werby23hovsg} have already constructed hierarchical scene graphs. However, they struggle to represent everyday dynamic environments due to two key challenges. First, their graph structures fail to capture the relationships between static objects and frequently moved everyday items. Second, frequently used daily objects often have changing positions and repetitive semantics, making efficient updates more difficult. 
Considering that frequently used objects are typically carried by static ones,  this article distinguishes  between static carriers and the objects they carry. We leverage LLM and VLM to identify these static carriers and construct an open-vocabulary carrier-relationship scene graph (CRSG), which effectively represents the carrying relationships between objects. In the process of robot navigation, new observations are matched with carrier objects in the carrier layer. For the observed carrier objects, dynamic updates are performed to reflect any changes in the everyday objects they carry.

Based on the CRSG, we designed an object-oriented navigation strategy, modeling the object search process as a Markov Decision Process (MDP) \cite{garcia2013markov}. At each navigation step, the robot decides to navigate toward candidate target objects or unexplored carrier objects based on visual-language feature similarities and the commonsense knowledge from the LLM, until the target is found.

In summary, our contributions are as follows:
\begin{itemize}
\item We present an adaptable carrier relationship scene graph (CRSG) that primarily describes the dynamic carrier and carried relationships between objects.
\item We design a navigation strategy based on the CRSG, utilizing visual-language features and commonsense knowledge from the LLM to inform decision-making.
\item Extensive qualitative and quantitative experiments demonstrate that our method effectively navigates to long sequences of moved objects, and the effectiveness of updating CRSG has been validated. Additionally, we deployed and tested the algorithm on a real robot, confirming its practicality.
\end{itemize}

\section{Related Work} \label{related work}

\subsection{Open Vocabulary Mapping}
With the advent of vision-language models like CLIP\cite{radford2021learning} and its variants, scene mapping with different representations has moved beyond the limitations of fixed classes\cite{deng2022hd,deng2022s}, and expanded to open-vocabulary \cite{jatavallabhula2023conceptfusion, peng2023openscene, lu2023ovir, gu2024conceptgraphs, deng2024opengraph,deng2024openobj}. Clip-fields\cite{shafiullah2022clip} integrates CLIP and SBERT features into the neural implicit map, enabling open-vocabulary map queries and navigation for robots. VLMap\cite{huang2023visual} and IVLMap\cite{huang2024ivlmap} project CLIP features top-down onto a 2D grid to enable zero-shot (instance) vision-language navigation. ConceptGraph\cite{gu2024conceptgraphs} and Hovsg\cite{werby23hovsg} constructed  instance-level point cloud maps with CLIP features embedded and scene graph representing certain relationships between objects, which facilitates more detailed and precise object retrieval.

These maps provide crucial support for applications such as open-vocabulary object queries, scene understanding, and robot navigation. However, they do not have dynamic update capabilities. We construct a carrier-relationship scene graph (CRSG) that describes the dynamic carrier and carried relationships between objects, and continuously updating the CRSG during navigation. This helps achieve more efficient navigation of everyday objects.

\subsection{Object Navigation}
Object navigation\cite{chang2020semantic, ye2021auxiliary,chaplot2020object,luo2022stubborn,ramakrishnan2022poni,zhang20233d, rajvanshi2024saynav, dorbala2023can, chen2023not,zhou2023esc,yokoyama2024vlfm}, as one of the key tasks in the field of embodied intelligence, primarily involves navigating to a specified semantic or instance location within a scene. \cite{chang2020semantic, ye2021auxiliary,chaplot2020object,luo2022stubborn,ramakrishnan2022poni,zhang20233d}  mainly perform object navigation within the close-classes.
\cite{gu2024conceptgraphs, huang2024ivlmap, werby23hovsg} constructed an offline instance map of the scene, enabling zero-shot instance navigation. \cite{dorbala2023can, chen2023not, zhou2023esc,yokoyama2024vlfm}  perform open-vocabulary object navigation using the frontier exploration method. However, they\cite{gu2024conceptgraphs, huang2024ivlmap, werby23hovsg,dorbala2023can, chen2023not, zhou2023esc,yokoyama2024vlfm} cannot capture changes in the positions of frequently used objects, or the addition and removal of instances in the scene. \cite{rajvanshi2024saynav} involves object navigation within close-classes and similarly does not involve scene updates. We have built a dynamic open vocabulary CRSG that not only supports  semantic object navigation but also enables efficient navigation to everyday instances (such as a red cup) that are spatially variable and subject to semantic interference. The approach most similar to ours is GOAT\cite{chang2023goat}, which also implements memory capabilities for the latest scene and supports multi-type and multi-modal navigation command inputs. However, for navigating to a displaced everyday object, we have designed a  navigation strategy based on CRSG, while GOAT selects the closest unexplored region to navigate to the object.

\begin{figure*}[ht]
\centering
{\includegraphics[width =17cm]{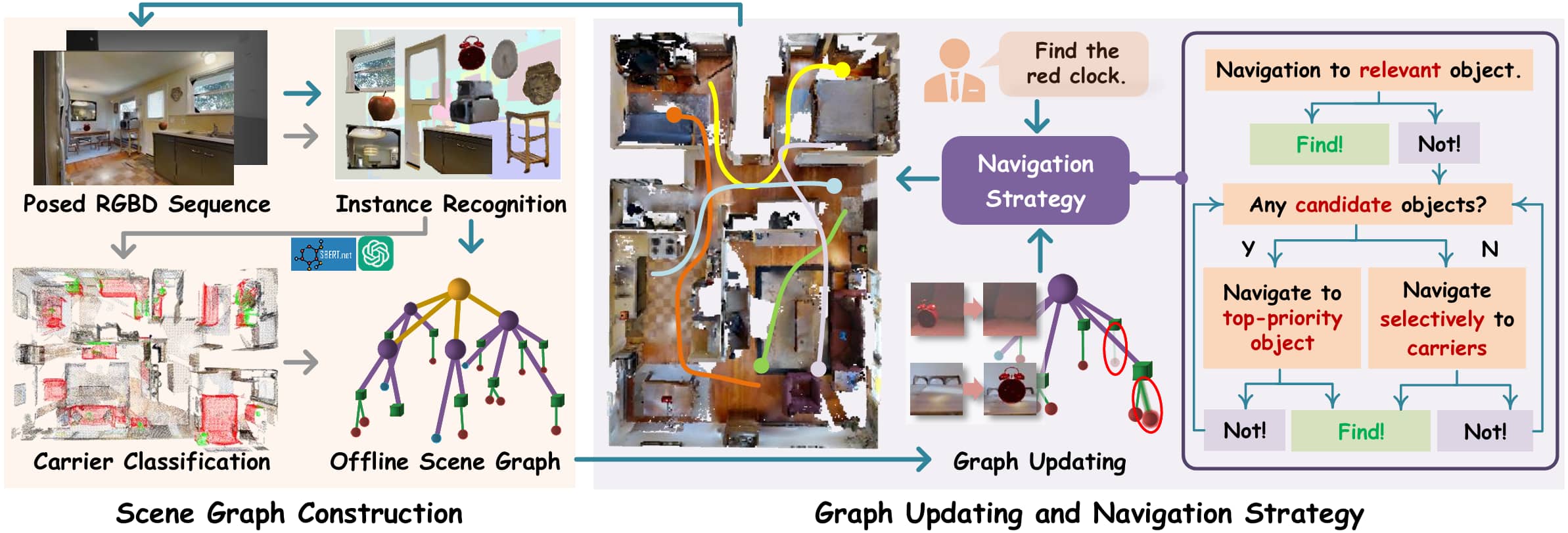}}
\caption{
The OpenObject-NAV system framework consists of two main modules. The Scene Graph Construction module focuses on constructing the carrier-relationship scene graph. The Graph Updating and Navigation Strategy module is responsible for executing cognitive navigation based on user instructions, following the proposed navigation strategy, while updating the scene graph in the process.
}
\setlength{\belowcaptionskip}{-0.2cm}
\label{system-framework}
\end{figure*} 

\textcolor{black}{\section{Method} \label{Systematic Framework}}

\subsection{Problem Definition}
In a daily environment, when given a navigation command, the robot queries the CRSG to determine the navigation endpoint and proceeds to the specified destination. If the target is a daily item (e.g., a cup) that is being carried, the robot evaluates whether the item remains in its original location based on the current observations. If not, the robot initiates a strategic exploration process. We define this challenge as a \textbf{displaced object exploration and navigation task} within an everyday setting. An overview of the system framework is provided in Fig.
\ref{system-framework}.
The cost function is defined as follows:

\begin{equation}
\label{cost_func}
P\_L= \sum_{t=1}^{T} \textit{Length}(L_t,L_{t+1})
\end{equation}

Let $L_t$ represent the position of the exploration target at step 
$t$, and $\textit{Length}(L_t,L_{t+1})$ denote the shortest path between $L_t$ and $L_{t+1}$, calculated by using path planning algorithms. Additionally, $T$ represents the number of exploration attempts the robot makes to navigate to the target object. \textcolor{black}{We aim to minimize 
$P\_L$ in Eq. \eqref{cost_func}.}

\subsection{Carrier-Relationship Scene Graph (CRSG)}
We first construct an open-vocabulary instance map $\mathcal{M}$ using the pre-collected RGB-D data of the scene. Unlike ConceptGraph\cite{gu2024conceptgraphs}, each instance object $\textit{\textbf{O}}_i \in \textit{\textbf{O}}$ (\textit{\textbf{O}} is the set of all objects) not only contains a CLIP feature $\textit{\textbf{V\_F}}_i$ but also stores a caption description list $\textit{\textbf{cap}}_i$ generated from the Tokenize Anything model \cite{pan2023tokenize} and text feature $\textit{\textbf{T\_F}}_i$ encoded with SBERT model \cite{reimers2019sentence}. A Carrier-Relationship Scene Graph (CRSG) $S\_G$ is then constructed below.

\textbf{{Building and Room layer:}} Existing works, such as \cite{hughes2022hydra, werby23hovsg}, propose various methods for room segmentation. We select specific point clouds from $\mathcal{M}$ and project them onto the $x-y$ plane: wall point clouds within a certain height range from the ground, and ``\textit{door}'' point clouds filtered using captions and text features. 
Next, we apply the line-search method, inspired by \cite{souxian}, to identify closed contours for each room, assigning objects within these contours to their corresponding rooms. The combined room layers constitute the building layer.

\textbf{{Carrier layer:}} We calculate the similarity between the text features $\textit{\textbf{T\_F}}_i$ of each object $\textit{\textbf{O}}_i$ and the SBERT-encoded text feature $\tilde{\textit{\textbf{T}}}$ for ``\textit{furniture for holding objects}''. The mathematical expression for this is as follows.

\begin{equation}
\label{sbert_sim}
sim(\textit{\textbf{T\_F}}_i,\tilde{\textit{\textbf{T}}}) = \frac{\textit{\textbf{T\_F}}_i \cdot
 \tilde{\textit{\textbf{T}}}}{||\textit{\textbf{T\_F}}_i|| \ ||\tilde{\textit{\textbf{T}}}||}
\end{equation}

Next, we select the set of objects $\tilde{\textit{\textbf{O}}} \subseteq \textit{\textbf{O}}$ with a similarity score exceeding a specified threshold $\sigma$, as shown below.

\begin{equation}
\label{sbert_thre}
 \tilde{\textit{\textbf{O}}} = \{ ({\textit{\textbf{O}}}_i,\textit{\textbf{cap}}_i,\textit{\textbf{T\_F}}_i) | sim(\textit{\textbf{T\_F}}_i,\tilde{\textit{\textbf{T}}}) > \sigma\}
\end{equation}

Next, we extract the three most frequent captions for each $\textit{\textbf{cap}}_i$ in $\tilde{\textit{\textbf{O}}}$, input them into a LLM (GPT-4o for test), and use a specific prompt to identify potential carrier-type objects, denoted as $\tilde{\textit{\textbf{O}}}_1 \subseteq \tilde{\textit{\textbf{O}}}$.

Finally, we select the final set of carrier-layer objects, denoted as $\bar{\textit{\textbf{O}}} 
\subseteq \tilde{\textit{\textbf{O}}}_1$, based on criteria such as the objects' geometric dimensions exceeding a certain size and their contact with the ground, as shown below.

\begin{equation}
\label{final_carrier}
 \bar{\textit{\textbf{O}}} \ = \ \textit{\textbf{f}} \ (\tilde{\textit{\textbf{O}}}_1)
\end{equation}

\textbf{{Object layer:}} For any non-carrier-layer object $\textit{\textbf{O}}_i \in (\textit{\textbf{O}}-\bar{\textit{\textbf{O}}})$, \textcolor{black}{we determine whether $\textit{\textbf{O}}_i$ is carried by a carrier-layer object $\textit{\textbf{O}}_j \in \bar{\textit{\textbf{O}}}$ based on $\textit{\textbf{O}}_i$'s dimensions, the closest distance, and the spatial overlap relationship in the x-y-z directions between $\textit{\textbf{O}}_i$ and $\textit{\textbf{O}}_j$ (exceeding  a certain overlap rate).} $\textbf{\textit{h}}(\textit{\textbf{O}}_j, \textit{\textbf{O}}_i)$ is defined to encapsulate the consideration of the aforementioned factors, where 
$\textbf{\textit{h}}(\textit{\textbf{O}}_j, \textit{\textbf{O}}_i)=1$
if all the conditions are satisfied. For any $\textit{\textbf{O}}_j \in \bar{\textit{\textbf{O}}}$, we define the set of objects $\textbf{\textit{C}}(\textit{\textbf{O}}_j)$ carried by $\textit{\textbf{O}}_j$
 as follows:


 \begin{equation}
\label{smalls_on_carrier}
 \textbf{\textit{C}}(\textit{\textbf{O}}_j) = \{\textit{\textbf{O}}_i| \textbf{\textit{h}}(\textit{\textbf{O}}_j, \textit{\textbf{O}}_i)=1, \textit{\textbf{O}}_i \in  (\textit{\textbf{O}}-\bar{\textit{\textbf{O}}}) \}
\end{equation}

\subsection{Navigation Strategy for a Displaced Object}
Let the \textbf{input navigation command} for the target object be either a  $\textbf{\textit{text}}$, or an $\textbf{\textit{image}}$. $\textbf{\textit{text}}$ or $\textbf{\textit{image}}$ is encoded using the SBERT or CLIP model, respectively. The resulting feature is then compared with the SBERT or CLIP features of each object in the CRSG $S\_G$ using  cosine similarity, similar to Eq. \eqref{sbert_sim}. The object with the highest similarity score is selected as the target object, $\textbf{\textit{O}}_{target}$.

We model the exploration of a displaced object as a fixed-policy Markov decision process (MDP) below.

\textbf{state space $S$:} In the current step $t$, we define:

\textbf{1.} the robot's pose $L_t \in \mathcal{L}$, 

\textbf{2.} the set of unexplored carrier-layer objects $CR_t \in \mathcal{CR}$, 

\textbf{3.} the set of candidate target objects \textit{on the unexplored carrier-layer objects} $CT_t \in \mathcal{CT}$,

\textbf{4.} the flag of finding the target or not $F_t \in \{0,1\}$. ( $\mathcal{L}$, $\mathcal{CR}$ and $\mathcal{CT}$ denote the value set of $L_t$, $CR_t$ and $CT_t$ respectively.) 

The state variable $S_t$ is defined in \eqref{state_variable}.

 \begin{equation}
\label{state_variable}
 S_t = (L_t, CR_t, CT_t, F_t) \in S
\end{equation}

In the initial state $S_0 = (L_0, CR_0, CT_0, F_0)$, $L_0$ is the initial position of the robot, $CR_0 = \bar{\textit{\textbf{O}}}$, and $CT_0=\textbf{\textit{O}}_{target}$.

\textbf{action space $A$:}
\begin{equation}
\label{action space}
 A = \{Stop, Explore(cr), Goto(ct) \ | \ cr \in CR_t, ct \in CT_t  \}
\end{equation}

$Stop$ indicates that the task is completed or all carrier-layer objects have been explored. $Explore(cr)$ and $Goto(ct)$ represent exploring the carrier-layer object $cr \in CR_t$ and navigating to the location of $ct \in CT_t$, respectively.

The robot selects the next action $a_t \in A$ based on the current state $S_t$ according to a specific policy $\pi(\cdot)$ in \eqref{action_policy}.

\begin{equation}
\label{action_policy}
 a_t = \pi(S_t)
\end{equation}

\textbf{policy $\pi(\cdot)$:}
Given current state $S_t = (L_t, CR_t, CT_t, F_t)$,   

\textbf{1.} if $F_t = 1$ or $CR_t=\emptyset$, then $a_t = Stop$. 

\textbf{2.} If $F_t = 0$ and $CT_t \ne \emptyset$,  we prioritize and select a candidate object to proceed with. Specifically, let $CT_t=\{ \textbf{\textit{O}}_{t1},..., \textbf{\textit{O}}_{ti} \}$. Some additional variables are stored: the SBERT similarities $SS_t=\{ \textbf{\textit{ss}}_{t1},..., \textbf{\textit{ss}}_{ti} \}$ between $CT_t$ and $\textbf{\textit{O}}_{target}$, the distances $D_t=\{ \textbf{\textit{d}}_{t1},..., \textbf{\textit{d}}_{ti} \}$ between $L_t$ and $CT_t$, and the average depth values $\tilde{D}_t=\{ \tilde{\textbf{\textit{d}}}_{t1},..., \tilde{\textbf{\textit{d}}}_{ti} \}$ when $CT_t$ are observed by the robot's camera. The priority rating of any $\textbf{\textit{O}}_{tj} \in CT_t$ corresponding to $\textit{\textbf{ss}}_{tj}$, $\textit{\textbf{d}}_{tj}$ and $\tilde{\textbf{\textit{d}}}_{tj}$, is evaluated as follows. 
The parameters in \eqref{candidate_priority} are set as $\omega_1=3$, $\alpha=0.1$ and $\omega_2=1$ in the experiments.

\begin{equation}
\label{candidate_priority}
 P\_R(\textbf{\textit{O}}_{tj}) = \frac{\omega_1 \cdot \textit{\textbf{ss}}_{tj} \cdot \exp(-\alpha\tilde{\textbf{\textit{d}}}_{tj})}{1+\omega_2 \cdot \textbf{\textit{d}}_{tj}}
\end{equation}
Where  $\textit{\textbf{ss}}_{tj}$ is positively correlated with $P\_R(\textbf{\textit{O}}_{tj})$, as we assume that a larger $\textit{\textbf{ss}}_{tj}$ indicates a higher likelihood that the candidate is the target. Moreover,
$\tilde{\textbf{\textit{d}}}_{tj}$ is negatively correlated with $P\_R(\textbf{\textit{O}}_{tj})$, based on the assumption that the accuracy of the front-end detection model decreases as $\tilde{\textbf{\textit{d}}}_{tj}$ increases. Therefore, $\exp(-\alpha\tilde{\textbf{\textit{d}}}_{tj})$ is considered the confidence level of $\textit{\textbf{ss}}_{tj}$. The robot will navigate to the location of the object with the maximum $P\_R$ and explore for $\textbf{\textit{O}}_{target}$.

\textbf{3.} If $F_t = 0$, $CT_t = \emptyset$ and $CR_t \ne \emptyset$, the LLM selects one of the carrier objects $cr_{k} \in CR_t$ and the robot executes the action $a_t=Explore(cr_{k})$. Specifically, the captions for each carrier object in $CR_t$ are extracted and provided as input to the LLM, along with the image or caption of the target object. Leveraging the LLM's commonsense understanding of object-carrier relationships (e.g., ``\textit{a cup is unlikely to be placed on a toilet}''), the LLM identifies the carrier object where the target object is most likely to be found. 

\textbf{state transition process:} 
 If $a_t =Explore(cr)$  (where $cr \in CR_t$) or  $a_t =Goto(ct)$  (where $ct \in CT_t$),  then  during the robot's movement, let $CR_{observed}$ represent the set of carrier objects observed within a small radius $r$ that have no candidate targets on them (based on the latest environmental observations), and let $CT_{new}$ represent the set of new target candidates found on unexplored carrier objects. Since some candidates in $CT_t$ may be carried by objects in $CR_{observed}$, $CT_t$ is updated to ${CT_t}^*$ after these candidates are removed. Specifically, the candidates in $CT_{new}$ are those for which the SBERT feature similarities with the target exceed a threshold $\sigma_1$. Additionally, the similarities between the target $\textbf{\textit{O}}_{target}$ and the objects carried in $CR_{observed}$ don't exceed $\sigma_1$.

\textbf{1.} if  $a_t=Explore(cr)$, $CR_{t+1}$ and $CT_{t+1}$ are updated as:

\begin{subequations}
\begin{align}
\label{CR_ex_t+1_}
 CR_{t+1} &= CR_t \ \backslash \ (\{cr\} \cup CR_{observed})  \\
 \label{CT_ex_t+1_}
 CT_{t+1} &= {CT_t}^*  \cup  CT_{new}
\end{align}
\end{subequations}

\textcolor{black}{\textbf{2.}} if \ $a_t=Goto(ct)$, $CR_{t+1}$ and $CT_{t+1}$ are updated as:
\begin{subequations}
\begin{align}
\label{goto_CR_t+1}
 CR_{t+1} &= CR_t \ \backslash \ (\{cr_1\} \cup CR_{observed}) , \ ct \in  \textbf{\textit{C}}(cr_1)
 \\
 \label{goto_CT_t+1_}
 CT_{t+1} &= {CT_t}^*  \cup  CT_{new} \ \backslash \ \{ct\}
\end{align}
\end{subequations}

In either case,  the SBERT feature similarities between $\textbf{\textit{O}}_{target}$ and the objects carried by $cr$ or $cr_1$ are calculated. If the input command is an image, an LLM-based image comparison is also performed. If the combined score from the LLM's image comparison and the SBERT text similarity for a carried object exceeds $\sigma_2$, then $F_{t+1}=1$, and the task is marked as complete. 

\subsection{CRSG Adaptation}

\textbf{Matching carrier objects.} As the robot navigates, it periodically captures RGB and depth images from the environment. The RGB images are processed through CropFormer\cite{qi2022high}, Tokenize Anything model \cite{pan2023tokenize}, CLIP\cite{radford2021learning} and SBERT\cite{reimers2019sentence} to obtain instance masks,  captions, encoded CLIP features and encoded SBERT features, respectively.  For newly observed objects, the robot compares them with the carrier objects in $S_G$ to identify the observed carrier objects $\textit{\textbf{O}}_{match}^{cr}$. The primary aspects of comparison include the object's size, the distance between their center positions, and the similarity scores based on CLIP and SBERT features.

\textbf{The Addition or Removal of Carried Objects.} For currently observed instances, $\textbf{\textit{h($\cdot,\cdot$)}}$ in Eq. \eqref{smalls_on_carrier} is used to determine whether they are being carried by $\textit{\textbf{O}}_{match}^{cr}$. Let the set of carried objects in the new observations be defined as $\textit{\textbf{O}}^{crd}$. The previously carried objects on $\textit{\textbf{O}}_{match}^{cr}$ are then compared with $\textit{\textbf{O}}^{crd}$. The criteria also include the object's size, the distance between center positions, and the SBERT feature similarity score. After the comparison, the carried objects on $\textit{\textbf{O}}_{match}^{cr}$ are updated accordingly: they are either added, removed, or left unchanged.

\section{Experimental Results}\label{experiment}
We aim to answer the following research questions: 

1. Does the carrier-relationship scene graph (CRSG) improve the accuracy of instance object queries (Sec. \ref{Offline_Object_Query})?

2. Does the dynamic update of the CRSG contribute to more efficient instance navigation (Sec. \ref{Long-sequence}, \ref{Ablation})? 

\textbf{Metrics.}
We report Success Rate(SR) and Success weighted by inverse Path Length (SPL)\cite{anderson2018evaluation}. SPL measures the efficiency of an robot's path by comparing it to the shortest route from the starting point to the target object instance. If the robot fails to reach the target, the SPL score is zero. Otherwise, the score is calculated as the ratio of the shortest path length to the robot's actual path length, with higher values indicating better performance.

\begin{table}[]
\centering
\caption{Success rate of object Query on an outdated offline map}
\label{tab1}
\resizebox{0.85\linewidth}{!}{
\begin{tabular}{ccccc}
\hline
Method & scene\_1& scene\_2& scene\_3&SR\\ \hline
 VLMap& 7/14& 8/19& 7/17&44\%\\
ConceptGraph& 9/14& 10/19&12/17&62\%\\
Ours& \textbf{13/14}& \textbf{15/19}&\textbf{15/17}&\textbf{86\%}\\
\hline
\end{tabular}
}\end{table}

\begin{figure*}[ht]
\centering
{\includegraphics[width =14.5cm]{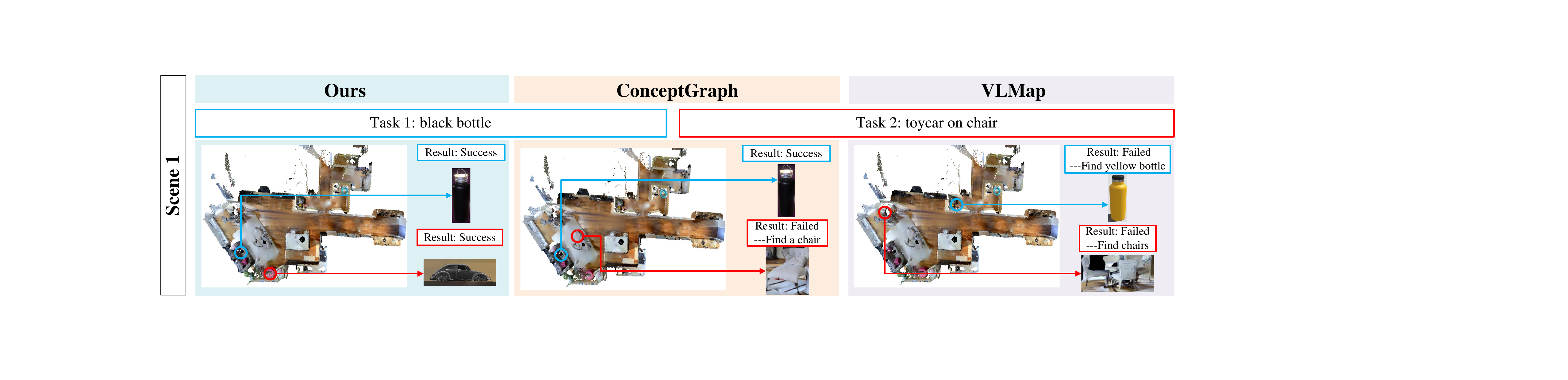}}
\caption{
Static Object Query Experiment: Comparison of Target Object Query Results on the Offline Map.
}
\setlength{\belowcaptionskip}{-0.2cm}
\label{expe1}
\end{figure*}

\begin{figure*}[ht]
\centering
{\includegraphics[width =15cm]{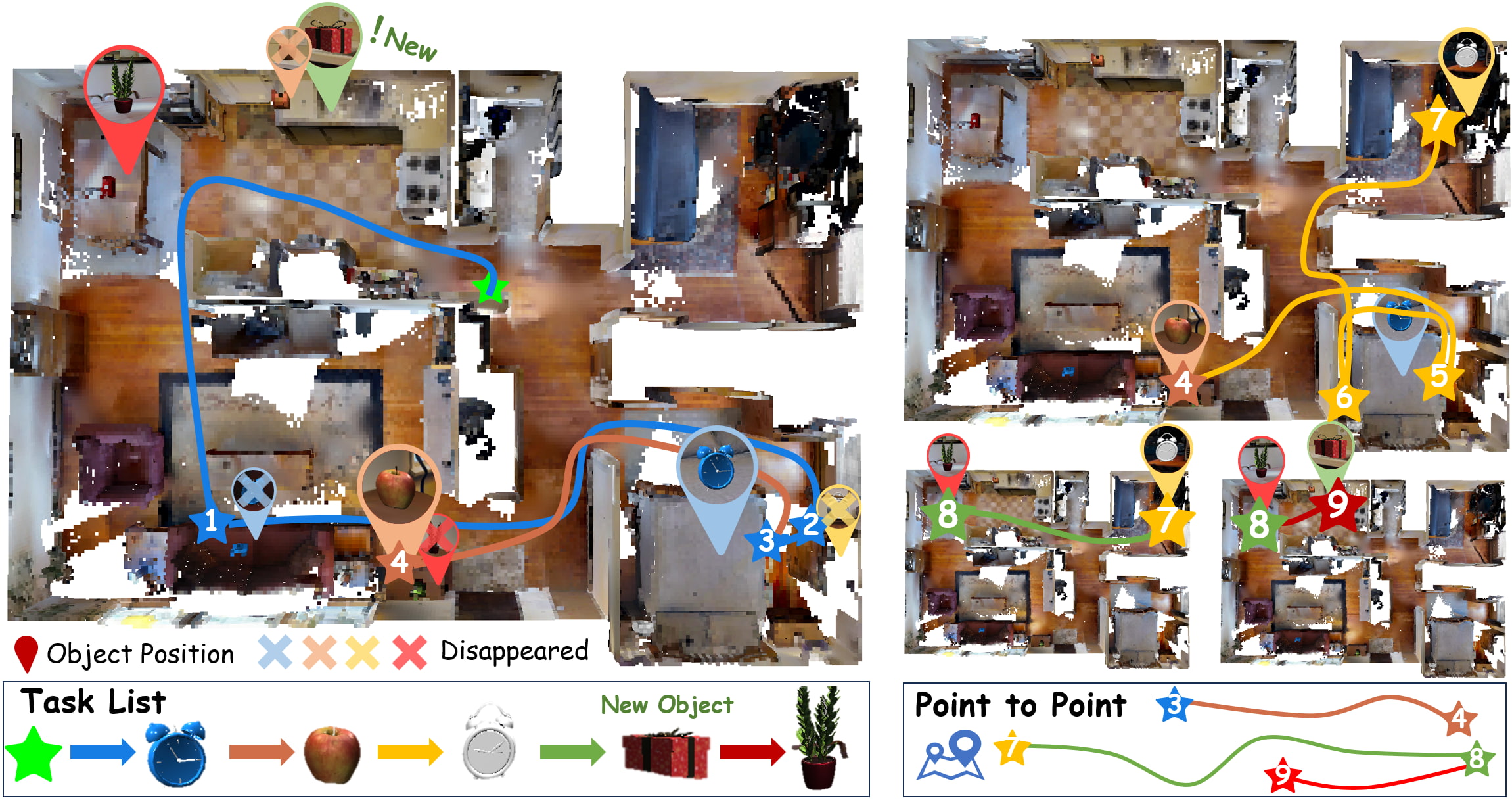}}
\caption{
The visualization of a long-sequence instance navigation result in scene 2 is shown, where "Point to Point" represents the shortest path navigation.
}
\setlength{\belowcaptionskip}{-0.2cm}
\label{expe2-show}
\end{figure*}

\subsection{Object Query on the Outdated Offline Map}
\label{Offline_Object_Query}
We compare the object query accuracy of scene representations with those of VLMap\cite{huang2023visual} and ConceptGraph\cite{gu2024conceptgraphs}. A total of 50 queries with different types of navigation instructions (semantic, instance and requirement-driven) were conducted across 3 scenes in Gibson\cite{xia2018gibson}. The experimental results are presented in Tab. \ref{tab1}, where our object query success rate averages 86\% and is the highest in all three scenarios. Because in instance queries like ``\textit{a cup on the table}'', the CRSG we constructed records the carrying relationship between \textit{cup} and \textit{table}, thus allowing for precisely locating the instance. In contrast, VLMap\cite{huang2023visual} and ConceptGraph\cite{gu2024conceptgraphs} may find the \textit{table} instead of the \textit{cup}. Besides, we additionally incorporates text features of caption descriptions for each object in CRSG, demonstrates superior performance in differentiating between similar objects, such as \textit{black cup} and \textit{white cup}. Meanwhile, VLMap projects the CLIP features from 3D space to a 2D grid, and its queryable semantic types are limited, resulting in inferior performance. We illustrate partial query results of different methods  in Fig. \ref{expe1}, and the results demonstrate that our method performs better in distinguishing between objects of the same category and in querying carried instances.

\begin{table*}[htbp]
\centering
\caption{SR and Tasks\_SR(\textit{i}) in different scenes for a series of long-sequence frequently used daily items navigation tasks}
\label{all_sr}
\large
\resizebox{0.9\textwidth}{!}{
\begin{tabular}{c|ccc|ccc|ccc}
\hline
\multirow{3}{*}{Object} & \multicolumn{9}{c}{SR (\%) / Tasks\_SR(\textit{i}) (\%)} \\
\cline{2-10}
 & \multicolumn{3}{c|}{scene\_1} & \multicolumn{3}{c|}{scene\_2} & \multicolumn{3}{c}{scene\_3} \\
\cline{2-10}
 & \textbf{ours-Text} & \textbf{ours-LLM} & \textbf{ours} & \textbf{ours-Text} & \textbf{ours-LLM} & \textbf{ours} & \textbf{ours-Text} & \textbf{ours-LLM} & \textbf{ours} \\
\hline
1 & 68.8 / 68.8 & 75.0 / 75.0 & \textbf{81.3 / 81.3} & 53.8 / 53.8 & 84.6 / 84.6 & \textbf{84.6 / 84.6} & 73.3 / 73.3 & 86.7 / 86.7 & \textbf{100 / 100} \\
2 & 56.3 / 37.5 & 68.8 / 50.0 & \textbf{75.0 / 56.3} & 76.9 / 53.8 & \textbf{100 / 84.6} & \textbf{100 / 84.6} & \textbf{100 } / 73.3 & 60.0 / 46.7 & \textbf{100 / 100} \\
3 & 68.8 / 31.3 & 56.3 / 25.0 & \textbf{75.0 / 31.3} & \textbf{84.6 / 46.2} & 76.9 / 61.5 & 76.9 / 61.5 & \textbf{100 } / 73.3 & 60.0 / 13.3 & \textbf{100 / 100} \\
4 & 81.3 / 18.8 & 87.5 / 25.0 & \textbf{100 / 31.3} & 69.2 / 23.1 & \textbf{92.3 / 53.8} & 84.6 / 46.2 & \textbf{100 / 73.3} & 66.7 / 6.7 & 73.3 / \textbf{ 73.3} \\
5 & 61.5 / 23.1 & 69.2 / 15.4 & \textbf{76.9 / 30.8} & 44.4 / 0 & \textbf{66.7 / 33.3} & \textbf{66.7 / 33.3} & \textbf{100 / 62.5} & 50.0 / 0 & \textbf{100 / 62.5} \\
\hline
\end{tabular}
}
\end{table*}

\subsection{Long-sequence Navigation Task for Frequently Used Everyday
Items}
\label{Long-sequence}
 
We conducted a series of long-sequence frequently used daily items navigation experiments (4 or 5 objects as a sequence) in three everyday scenarios in Gibson\cite{xia2018gibson} using the Habitat navigation simulator\cite{savva2019habitat}. In each scene, we placed a variety of frequently used items from different instances (like black cup, blue alarm clock, plastic bottle, game controller and so on), while offline constructing the CRSG for each scene. We set up multiple distractor objects of the same category (like black / white cup) to validate our ability to correctly navigate to a specific instance. We then randomly altered the positions of these items to simulate the variability in the locations of commonly used everyday objects. Next, the robot is instructed to sequentially navigate to these objects in each scene. 

\begin{figure}[t!]
\centering    
{
	\includegraphics[width=8.5cm]{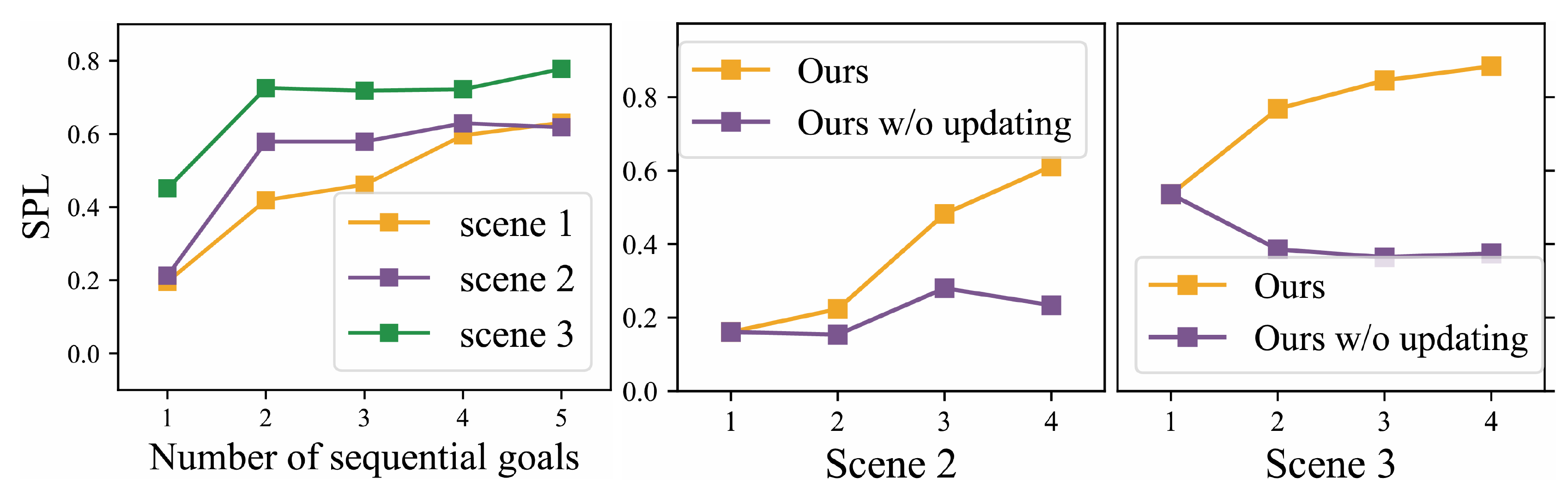} 

}
\caption{The \textbf{first} figure presents the SPL results in Sec. \ref{Long-sequence}, while the \textbf{second} and \textbf{third} figures show the results of the ablation experiments with and without CRSG updates in Sec. \ref{Ablation}.} %
\label{spl} 
\end{figure}

We present the navigation results for each scene in Tab. \ref{all_sr} for SR and Tasks\_SR(\textit{i}), and Fig. \ref{spl} for SPL. Tasks\_SR(\textit{i}) represents the success rate of correctly navigating to all 
\textit{i} objects. We also presented SR in Tab. \ref{all_sr} when using only SBERT feature similarity (\textbf{ours-Text}) and GPT-4o (\textbf{ours-LLM}) for image matching to determine targets. Tab. \ref{all_sr} shows that  \textbf{ours} achieves generally the highest SR and Tasks\_SR(\textit{i}). This indicates that considering both SBERT feature similarity and the image matching results from GPT-4o  contributes to navigating to the true-positive target. Fig. \ref{expe2-show} illustrates an example of long-sequence navigation, where the efficiency of navigating to the target significantly improves as the number of navigated objects increases. Additionally, as shown in Fig. \ref{spl}, the  SPL for the first object is noticeably lower, while the SPL for the remaining objects shows significant improvement. Since the process of navigating to the first object involves multiple explorations, it often results in a navigation path length that greatly exceeds the shortest path. During this period, the observed CRSG is updated, enabling the positions of other objects to be refreshed. As a result, when navigating to the remaining objects (including new instances), point-to-point navigation is primarily achieved, leading to a increase of SPL.

\begin{table}[]
\centering
\caption{Ablation Study: SPL of single object navigation}
\label{module-ablation}
\resizebox{\linewidth}{!}{
\begin{tabular}{cccc}
\hline
Metric & \textbf{only-carriers\_Random} & \textbf{only-carriers\_LLM} & \textbf{ours}\\ \hline
SPL &0.205 & 0.309 & \textbf{0.342} \\
\hline
\end{tabular}

}\end{table}

\subsection{Ablation Study}
\label{Ablation}
To further investigate the role of CRSG updates in efficient navigation to everyday objects, we conducted ablation experiments on one long-sequence navigation tasks in each of Scene 2 and Scene 3, comparing cases with and without CRSG updates. The results (in Fig. \ref{spl}, second and third figures) indicate that when CRSG is updated, the SPL gradually increases, while there is no improvement in SPL without CRSG updates. Thus, the updates to CRSG contribute to more efficient navigation to displaced everyday objects. 

We also conducted single daily object navigation experiments in three different  scenes to evaluate the impact of various modules in our navigation strategy on navigation efficiency. \textbf{only-carriers\_Random} represents navigating to a randomly selected carrier object for exploration, without considering candidate target objects. \textbf{only-carriers\_LLM} selects the next carrier object for exploration based on LLM's recommendations, building upon \textbf{only-carriers\_Random}. As shown in Tab. \ref{module-ablation}, our method achieves the highest SPL, followed by \textbf{only-carriers\_LLM}. This indicates that the strategy of navigating to candidate target objects and selecting the carrier objects to explore based on the commonsense knowledge of LLM contributes to more efficient navigation.

\begin{figure}[t!]
\centering    
{
	\includegraphics[width=8cm]{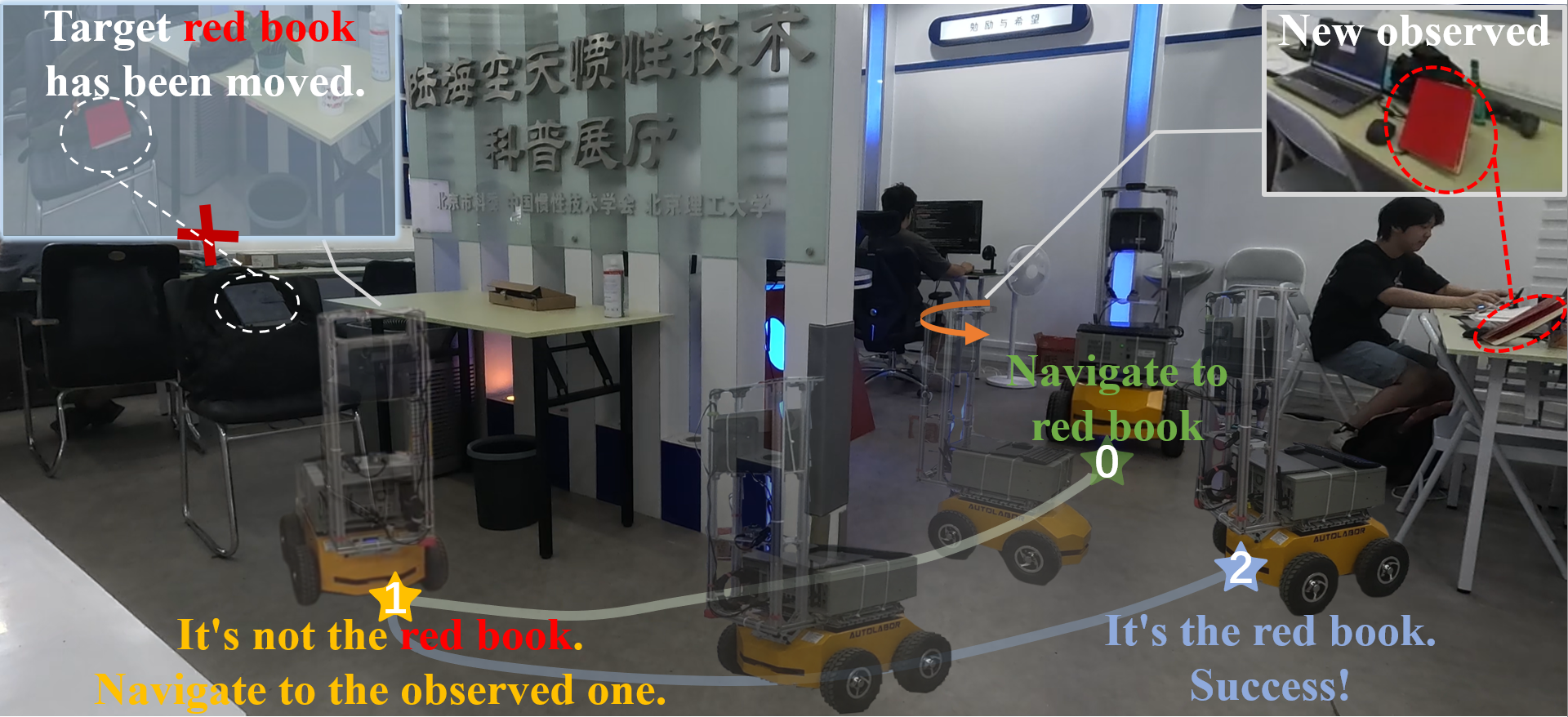} 

}
\caption{The robot queries the CRSG for the position of the \textbf{red book} at the \textbf{chair} and navigates there. It then discovers that the red book is not in its original location and rules out the interference of the \textbf{grey book}. Finally, it navigates to the \textbf{newly observed} position of the red book.}%
\label{real} 
\end{figure}

\subsection{Real-World Validation}
We validated our algorithm using an Autolabor robot in a real scene, equipped with an industrial computer featuring an NVIDIA GeForce RTX 3080. We equipped the robot with a Livox Mid 360 LiDAR and utilized the Cartographer SLAM algorithm\cite{xu2017research} to obtain  its global pose. Additionally, an Azure Kinect DK was mounted to capture RGB-D information. The robot successfully navigates to a displaced red book shown in Fig. \ref{real}.

\section{CONCLUSIONS} \label{CONCLUTIONS}
This paper has proposed an open-vocabulary navigation method for frequently used everyday items, leveraging a dynamic carrier-relationship scene graph (CRSG). Specifically, we first construct the CRSG to capture the dynamic relationships between carrier objects and the objects they carry. Next, a navigation strategy based on the CRSG is developed to navigate to frequently used items, modeling the object search process as a Markov Decision Process (MDP). At each navigation step, the CRSG is dynamically updated based on the robot's observations of the environment. The robot then decides whether to navigate toward candidate target objects or unexplored carrier objects, guided by visual-language feature similarities and commonsense knowledge from the LLM. Both simulations and physical experiments demonstrate  that our method efficiently navigates to objects that are subject to positional changes, even in the presence of distractors from the same category.  In the future, we plan to incorporate an online mapping module and  parallel processing  to improve exploration efficiency.

\bibliographystyle{IEEEICRA}
\bibliography{bib/ICRA}
\end{document}